# 11 TeraFLOPs per second photonic convolutional accelerator for deep learning optical neural networks

Xingyuan Xu,[1] Mengxi Tan,[1] Bill Corcoran,[2] Jiayang Wu,[1] Andreas Boes,[3] Thach G. Nguyen,[3] Sai T. Chu,[4] Brent E. Little,[5] Damien G. Hicks,[1,6] Roberto Morandotti,[7,8] Arnan Mitchell,[3] and David J. Moss [1,*]

[1] Optical Sciences Centre, Swinburne University of Technology, Hawthorn, VIC 3122, Australia

[2] Department of Electrical and Computer Systems Engineering, Monash University, Clayton, 3800 VIC, Australia

[3] School of Engineering, RMIT University, Melbourne, VIC 3001, Australia

[4] Department of Physics, City University of Hong Kong, Tat Chee Avenue, Hong Kong, China.

[5] Xi'an Institute of Optics and Precision Mechanics Precision Mechanics of CAS, Xi'an, China.

[6] Bioinformatics Division, Walter & Eliza Hall Institute of Medical Research, Parkville, Victoria 3052, Australia

[7] INRS-Énergie, Matériaux et Télécommunications, 1650 Boulevard Lionel-Boulet, Varennes, Québec, J3X 1S2, Canada.

[8] Institute of Fundamental and Frontier Sciences, University of Electronic Science and Technology of China, Chengdu 610054, PR China

*Corresponding author: dmoss@swin.edu.au

**Abstract:** Convolutional neural networks (CNNs), inspired by biological visual cortex systems, are a powerful category of artificial neural networks that can extract the hierarchical features of raw data to greatly reduce the network parametric complexity and enhance the predicting accuracy. They are of significant interest for machine learning tasks such as computer vision, speech recognition, playing board games and medical diagnosis [1-7]. Optical neural networks offer the promise of dramatically accelerating computing speed to overcome the inherent bandwidth bottleneck of electronics. Here, we demonstrate a universal optical vector convolutional accelerator operating beyond 10 Tera-FLOPS (floating point operations per second), generating convolutions of images of 250,000 pixels with 8-bit resolution for 10 kernels simultaneously — enough for facial image recognition. We then use the same hardware to sequentially form a deep optical CNN with ten output neurons, achieving successful recognition of full 10 digits with 900 pixel handwritten digit images with 88% accuracy. Our results are based on simultaneously interleaving temporal, wavelength and spatial dimensions enabled by an integrated microcomb source. This approach is scalable and trainable to much more complex networks for demanding applications such as unmanned vehicle and real-time video recognition.

Artificial neural networks (ANNs) are collections of nodes with weighted connections that, with proper feedback to adjust the network parameters, can "learn" and perform complex operations for face recognition, speech translation, playing board games and medical diagnosis [1-4]. While classic fully connected feedforward networks face challenges in processing extremely high-dimensional data, convolutional neural networks (CNNs), inspired by the (biological) behavior of the visual cortex system, can abstract the representations of input data in their raw form, and then predict their properties with both unprecedented accuracy and greatly reduced parametric complexity [5]. CNNs have been widely applied to computer vision, natural language processing and other areas [6, 7].

The capability of neural networks is dictated by the computing power of the underlying neuromorphic hardware. Optical neural networks (ONNs) [8-12] are promising candidates for next-generation neuromorphic computation, since they have the potential to overcome the bandwidth bottleneck of their electrical counterparts [6, 13-16] and achieve ultra-high computing speeds enabled by the >10 THz wide optical telecommunications band [8]. Operating in analog frameworks, they avoid the limitations imposed by the energy and time consumed during reading and storing data back and forth, known as the von Neumann bottleneck [13]. Significant progress has been made in highly parallel, high-speed and trainable ONNs [8-12, 17-21], including approaches that have the potential for full integration on a single photonic chip [8, 12], in turn offering an ultra-high computational density. However, there remains opportunities for significant improvements in ONNs. Processing large-scale data, as needed for practical real-life computer vision tasks, remains challenging for ONNs because they are primarily fully connected structures where their input scale is determined solely by hardware parallelism. This leads to tradeoffs between the network scale and footprint. Moreover, ONNs have not achieved the extreme computing speeds that analog photonics is capable of, given the very wide optical bandwidths that they can exploit.

Here, we demonstrate an optical convolution accelerator that operates beyond 10 Tera-FLOPS (floating point operations per second) and use it to process and compress large-scale data. Through interleaving wavelength, temporal, and spatial dimensions using an integrated Kerr frequency comb (or "microcomb" [22-31]), we achieve a vector computing speed as high as 11.322 Tera-FLOPS. We then use it to process images with a length of 250,000 pixels with ten convolution kernels at 3.8 TeraFLOPs. Our convolution accelerator is fully and dynamically reconfigurable, as well as scalable, so that it can serve as both a convolutional accelerator front-end to generate convolutions with multiple and simultaneous parallel kernels, as well as forming an optically deep CNN with fully connected neurons, without any change in hardware. We use the deep CNN to achieve successful recognition of the full ten digits (0-9) for handwritten images, achieving and accuracy of 88%. Our optical neural network represents a major step towards realizing monolithically integrated ONNs and is enabled by our use of an integrated microcomb chip. Moreover, our accelerator scheme is stand alone and universal — fully compatible with either electrical or optical interfaces. Hence, it can serve as a universal ultrahigh bandwidth data compressing front end for any neuromorphic hardware — either optical or electronic based — bringing massive-data machine learning for both real-time and ultrahigh bandwidth data within reach.

## Principle of Operation

Figure 1 shows the principle of operation for the photonic vector convolutional accelerator (VCA) which features high-speed electrical signal ports for data input and output. The input data vector **X** is encoded as the intensity of temporal symbols in a serial electrical waveform at a symbol rate $1/\tau$ (baud), where $\tau$ is the symbol period. The convolution kernel is similarly represented by a weight vector **W** of length $R$ that is then encoded in the optical power of the microcomb lines through spectral shaping performed by a Waveshaper. The temporal waveform **X** is then multi-cast onto the kernel wavelength channels via electro-optical modulation, thus generating the replicas weighted by **W**. Next the optical waveform is transmitted through a dispersive delay with a delay step (between adjacent wavelength channels) equal to the symbol duration of **X**, effectively achieving time and wavelength interleaving. Finally, the delayed and weighted replicas are summed via high speed photodetection so that each time slot yields a convolution between **X** and **W** for a given convolution window, or receptive field. As such, the convolution window effectively slides at the modulation speed matching the baud rate of **X**. Each output symbol is the result of $R$ multiply-and-accumulate operations, with the computing speed given by $2R/\tau$ FLOPS. Since the speed of this process scales with both the baud rate and number of wavelengths, it can be dramatically boosted into the Tera-FLOP regime by using the massively parallel wavelength channels of the microcomb source. Moreover, the length of the input data **X** is theoretically unlimited so that the convolution accelerator can process data with an arbitrarily large scale—the only practical limitation being the capability of the external electronics.

We achieve the simultaneous convolution of multiple kernels in parallel simply by adding additional sub-bands of $R$ wavelengths for each additional kernel. Following multicasting and dispersive delay, the sub-bands (kernels) are demultiplexed and detected separately with high speed photodetectors, generating a separate electronic waveform for each kernel. The VCA is fully reconfigurable and scalable – the number and length of the kernels are arbitrary, limited only by the total number of wavelengths.

While the core convolutional accelerator system typically processes vectors, it can easily be adapted to operate on matrices for image processing. For optical processing of matrix operations, the matrix must first be flattened into a vector, and the precise way that this is performed determines both the sliding convolution window's stride and the equivalent matrix computing speed. Our flattening method sets the receptive field (convolution slot) to slide with a horizontal stride of unity (ie., every matrix input element has a corresponding convolution output) and a vertical stride that scales with the size of the convolutional kernel. The larger vertical stride effectively resulted in sub-sampling across the vertical direction of the raw input matrix, equivalent to a partial pooling function [32] in addition to convolution. This resulted in an effective reduction (or overhead) in matrix computing speed that scales inversely with the size of the kernel (eg., a 3x3 kernel yields an overhead (speed reduction) of a factor 3). While this can be alleviated by various means to produce convolutions with a symmetric stride and no speed overhead, this is actually not necessary for most applications (see Supplementary Materials).

Finally, this approach is highly flexible and reconfigurable without any change in hardware - we use same system for the convolutional accelerator for image processing as well as to form an optical deep learning CNN which we use to perform a separate series of experiments. The convolutional accelerator hardware forms both the input processing stage as well as the fully connected neuron layer of the CNN (see below). The system can achieve matrix multiplication by simply sampling one time slot of the output waveform, since the vector dot product is equivalent to the special convolution case where the two input vectors **X** and **W** have the same length.

## Experiment

**Matrix Convolutional Accelerator**

Figure 2 shows the experimental setup for the full matrix convolutional accelerator that we use to process a classic 500×500 face image. The system performs 10 simultaneous convolutions with ten 3×3 kernels to achieve distinctive image processing functions. The weight matrices for all kernels were flattened into a composite kernel vector **W** containing all 90 weights (10 kernels with 3x3=9 weights each), which were then encoded onto the optical power of 90 microcomb lines by an optical spectral shaper (Waveshaper), each kernel occupying its own frequency band of 9 wavelengths. The wavelength channels were supplied by a coherent soliton crystal microcomb (Fig. 3) via optical parametric oscillation in a single micro-ring resonator (MRR Fig. 3b) [22−31] with a radius of 592 μm [22, 23], corresponding to a spacing of ~ 48.9 GHz [31] with an optical bandwidth of ~ 36 nm for the 90 wavelengths across the telecommunications C-band (1540-1570 nm) (see Methods and Supplementary) [30].

The raw 500×500 input face image was flattened electronically into a vector **X** and encoded as the intensities of 250,000 temporal symbols with a resolution of 8 bits/symbol (limited by the electronic arbitrary waveform generator (AWG)), to form the electrical input waveform via a high-speed electrical digital-to-analog converter, at a data rate of 62.9 Giga Baud (time-slot $\tau$ =15.9 ps) (Fig. 4b). The waveform duration was 3.975μs for each image corresponding to a processing rate for all ten kernels of over 1/3.975μs, equivalent to 0.25 million of these ultra-large-scale images per second.

The input waveform **X** was then multi-cast onto the 90 shaped comb lines via electro-optical modulation, yielding replicas weighted by the kernel vector **W**. Following this, the waveform was then transmitted through a ~2.2 km length of standard single mode fibre having a dispersion of ~17ps/nm/km. The fibre length was carefully chosen to induce a relative temporal shift in the weighted replicas with a progressive delay step of 15.9 ps between adjacent wavelength channels. This delay exactly matched the duration of each input data symbol $\tau$, which effectively resulted in time and wavelength interleaving for all ten kernels.

The 90 wavelengths were then de-multiplexed into 10 sub-bands of 9 wavelengths, each sub-band corresponding to a kernel, and separately detected by 10 high speed photodetectors. The detection process effectively summed the aligned symbols of the replicas (the electrical output waveform of one of the kernels (*kernel 4*) is shown in Fig. 4c). The 10 electrical waveforms were converted into digital signals via ADCs and resampled so that each time slot of each of the waveforms corresponded to the dot product between one of the convolutional kernel matrices and the input image within a sliding window (i.e., receptive field). This effectively achieved convolutions between the 10 kernels and the raw input image. The resulting waveforms thus yielded the 10 feature maps (convolutional matrix outputs) containing the extracted hierarchical features of the input image (Fig. 4d, Supplementary Materials).

The convolutional vector accelerator makes full use of time, wavelength, and spatial multiplexing, where the convolution window effectively slides across the input vector **X** at a speed equal to the modulation baud-rate — 62.9 Giga Symbols/s. Each output symbol is the result of 9 (the length of each kernel) multiply-and-accumulate operations, thus the core vector computing speed (i.e., *throughput*) of each kernel is 2×9×62.9 = 1.13 Tera FLOPS. For ten kernels computed in parallel the overall computing speed of the VCA is therefore 1.13×10 =11.3 Tera FLOPS, or 11.321×8=90.568 tera-bits per second (Tb/s) (reduced slightly by the optical signal to noise ratio (OSNR)). This speed is over 500 times higher than the fastest speed of ONNs reported to date (see supplementary).

For the image processing matrix application demonstrated here, the convolution window had a vertical sliding stride of 3 (resulting from the 3×3 kernels), and so the effective matrix computing speed was 11.3/3=3.8 TeraFLOPs. Homogeneous strides operating at the full vector speed can be readily achieved by duplicating the system with parallel weight-and-delay paths (see Supplementary Materials), although we found that this was unnecessary. While the length of the input data processed here was 250,000 pixels, the convolution accelerator can process data with an arbitrarily large scale, the only practical limitation being the capability of the external electronics.

**Deep Learning Optical Convolutional Neural Network**

The convolutional accelerator architecture presented here is fully and dynamically reconfigurable and scalable with the same hardware system. We were thus able to use the accelerator to sequentially form both a frontend convolution processor as well as a fully connected layer, together yielding an optical deep CNN. We applied the CNN to the recognition of full 10 (0-9) handwritten digit images.

Figure 5 shows the overall principle of the optical deep CNN while Figure 6 shows the detailed experimental configuration. The convolutional layer performs the heaviest computing duty of the entire network, generally taking 55% to 90% of the total computing power, and operated as described in the previous section. The digit images – 30×30 matrices of grey-scale values with 8 bit resolution – were flattened into vectors and multiplexed in the time-domain at 11.9 Giga Baud (time-slot $\tau$ =84 ps). Three 5×5 kernels were used, requiring 75 microcomb lines (Fig. 7) and hence resulted in a vertical stride of 5. The dispersive delay was achieved with ~13 km of standard SMF to match the data baud-rate. The wavelengths were de-multiplexed into the three kernels which were detected by high speed photodetectors and then sampled and nonlinearly scaled with digital electronics to recover the extracted hierarchical feature maps of the input images. The feature maps were then pooled electronically and flattened into a vector $\mathbf{X}_{FC}$ (72×1= 6×4×3) per image that formed the input data to the fully connected layer.

The fully connected layer had ten neurons, each corresponding to one of the ten categories of handwritten digits from 0 to 9, with the synaptic weights represented by a 72×10 weight matrix $\mathbf{W}_{FC}^{(l)}$ (ie., ten 72×1 column vectors) for the $l$th neuron ($l \in [1, 10]$) – with the number of comb lines (72) matching the length of the flattened feature map vector $\mathbf{X}_{FC}$. The shaped optical spectrum at the $l$th port had an optical power distribution proportional to the weight vector $\mathbf{W}_{FC}^{(l)}$, thus serving as the equivalent optical input of the $l$th neuron. After being multicast onto the 72 wavelengths and progressively delayed, the optical signal was weighted and demultiplexed with a single Waveshaper into 10 spatial output ports — each corresponding to a neuron. Since this part of the network involved linear processing, the kernel wavelength weighting could be implemented either before the EO modulation or at a later stage just before photodetection. The advantage of the latter configuration is that both the demultiplexing and weighting can then be achieved with a single Waveshaper. Finally, the different node/neuron outputs were obtained by sampling the 73th symbol of the convolved results. The final output of the optical CNN was represented by the intensities of the output neurons (Fig. 8, see supplementary), where the highest intensity for each tested image corresponded to the predicted category. The peripheral systems, including signal sampling, nonlinear function and pooling, were implemented electronically with digital signal processing hardware, although some of these functions (e.g., pooling) can in principle be performed in the optical domain with the VCA. Supervised network training was performed offline electronically (see Supplementary Materials).

We experimentally tested fifty 8-bit 30 × 30 resolution images of the handwritten digit dataset with the deep optical CNN. The confusion matrix (Figure 8, see Supplementary for definition) shows an accuracy of 88% for the generated predictions, in contrast to 90% for the numerical results calculated on an electrical digital computer. The computing speed of the VCA component of the deep optical CNN was 2×75×11.9 =1.785 Tera FLOPS, or 14.3 Terabits/s. For the application to process the image matrices with 5×5 kernels, the convolutional layer had a matrix flattening overhead of 5, yielding an image computing speed of 1.785/5= 357 Giga FLOPS. The computing speed of the fully connected layer was 119.8 Giga-FLOPS (see Supplementary Materials). The waveform duration was 30×30×84ps=75.6ns for each image, and so the convolutional layer processed images at the rate of 1/75.6ns = 13.2 million handwritten digit images per second.

We note that handwritten digit recognition, although widely employed as a benchmark test in digital hardware, is still (for full 10 digit (0 - 9) recognition) beyond the capability of existing analog reconfigurable ONNs. Digit recognition requires a large number of physical parallel paths for fully-connected networks (e.g., a hidden layer with 10 neurons requires 9000 physical paths), which poses a huge challenge for current nanofabrication techniques. Our CNN represents the first reconfigurable and integrable ONN capable not only of performing high level complex tasks such as full handwritten digit recognition, but at ultrahigh TeraFLOP speeds.

## Discussion

This approach can be readily scaled in performance in terms of input data size, as well as network size and speed. The data size is limited in practice only by the memory of the electrical digital-to-analog converters, and so in principle it is possible to process 4K-resolution (4096×2160) images. By integrating 100 photonic convolution accelerators layers (still much less than the 65536 processors integrated in the Google TPU [15]), the optical CNN would be capable of solving much more difficult image recognition tasks at a vector computing speed of 100 × 11.3=1.130 Peta-FLOPS. Further, the optical CNN presented here supports online training, since the optical spectral shaper used to establish the synapses can be dynamically reconfigured with a response time of < 500 ms or even faster with integrated optical spectral shapers [33].

Although the current embodiment presented here had a non-trivial optical *latency* of 0.11 μs introduced by the propagation delay of the dispersive fibre spool, this did not affect the operational speed. Moreover, the latency of the delay function can be virtually eliminated (to < 200 ps) by using integrated highly dispersive devices such as photonic crystals or customized chirped Bragg gratings [34].

Finally, current nanofabrication techniques can enable significantly higher levels of integration of the convolutional accelerator. The micro-comb source itself is based on a CMOS compatible platform that is intrinsically designed for large-scale integration. Other components such as the optical spectral shaper, modulator, dispersive media, de-multiplexer and photodetector have all been realized in integrated (albeit simpler) forms [33-35].

# Conclusion

We demonstrate a universal optical convolutional accelerator operating at 11.3 Tera-FLOPS for vector processing, and use a matrix processing version to perform convolutions on face images with 250,000 8-bit resolution pixels. We then use it to sequentially form an optical deep learning CNN to achieve successful recognition of handwritten digit images. Our network is capable of recognizing and processing large-scale data and images at ultra-high computing speeds for real-time massive-data machine learning tasks, such as identifying faces in cameras or pathology identification in clinical scanning applications [36, 37].

**References**

1. LeCun, Y., Bengio, Y. & Hinton, G. Deep learning. Nature **521**, 436–444 (2015).
2. Schalkoff, R. J. Pattern recognition. Wiley Encyclopedia of Computer Science and Engineering (2007).
3. Mnih, V. et al. Human-level control through deep reinforcement learning. Nature **518**, 529–533 (2015).
4. Silver, D. et al. Mastering the game of Go without human knowledge. Nature **550**, 354 (2017).
5. Krizhevsky, A., Sutskever, I. & Hinton, G. E. ImageNet Classification with Deep Convolutional Neural Networks. Commun Acm 60, 84-90 (2017).
6. Yao, P. et al. Fully hardware-implemented memristor convolutional neural network. Nature **577**, 641–646 (2020).
7. Lawrence, S., Giles, C. L., Tsoi, A. C. & Back, A. D. Face recognition: A convolutional neural-network approach. IEEE transactions on neural networks 8, 98-113 (1997).
8. Shen, Y. et al. Deep learning with coherent nanophotonic circuits. Nature Photonics **11**, 441 (2018).
9. Larger, L. et al. High-speed photonic reservoir computing using a time-delay-based architecture: Million words per second classification. Physical Review X **7**, 011015 (2017).
10. Peng, H., Nahmias, M. A., Lima, T. F. d., Tait, A. N. & Shastri, B. J. Neuromorphic Photonic Integrated Circuits. IEEE Journal of Selected Topics in Quantum Electronics **24**, 1-15, doi:10.1109/JSTQE.2018.2840448 (2018).
11. Lin, X. et al. All-optical machine learning using diffractive deep neural networks. Science **361**, 1004-1008 (2018).
12. Feldmann, J., Youngblood, N., Wright, C. D., Bhaskaran, H. & Pernice, W. H. P. All-optical spiking neurosynaptic networks with self-learning capabilities. Nature 569, 208-214, doi:10.1038/s41586-019-1157-8 (2019).
13. Ambrogio, S. *et al.* Equivalent-accuracy accelerated neural-network training using analogue memory. *Nature* **558**, 60 (2018).
14. Esser, S. K. et al. Convolutional networks for fast, energy-efficient neuromorphic computing. Proceedings of the National Academy of Sciences **113**, 11441, doi:10.1073/pnas.1604850113 (2016).
15. Graves, A. et al. Hybrid computing using a neural network with dynamic external memory. Nature **538**, 471–476 (2016).
16. Miller, D. A. B. Attojoule Optoelectronics for Low-Energy Information Processing and Communications. Journal of Lightwave Technology **35**, 346-396 (2017).
17. Appeltant, L. et al. Information processing using a single dynamical node as complex system. Nature Communications **2**, 468 (2011).

18. Chang, J., Sitzmann, V., Dun, X., Heidrich, W. & Wetzstein, G. Hybrid optical-electronic convolutional neural networks with optimized diffractive optics for image classification. Scientific Reports 8 (2018).
19. Vandoorne, K. et al. Experimental demonstration of reservoir computing on a silicon photonics chip. Nature Communications **5**, 3541 (2014).
20. Brunner, D., Soriano, M. C., Mirasso, C. R. & Fischer, I. Parallel photonic information processing at gigabyte per second data rates using transient states. Nature communications 4, 1364 (2013).
21. Tait, A. N., Chang, J., Shastri, B. J., Nahmias, M. A. & Prucnal, P. R. Demonstration of WDM weighted addition for principal component analysis. Optics Express **23**, 12758-12765 (2015).
22. Pasquazi, A. et al. Micro-combs: a novel generation of optical sources. Physics Reports **729**, 1-81 (2018).
23. Moss, D. J., Morandotti, R., Gaeta, A. L. & Lipson, M. New CMOS-compatible platforms based on silicon nitride and Hydex for nonlinear optics. Nature photonics **7**, 597 (2013).
24. Kippenberg, T. J., Gaeta, A. L., Lipson, M. & Gorodetsky, M. L. Dissipative Kerr solitons in optical microresonators. Science **361**, 567 (2018).
25. Savchenkov, A. A. et al. Tunable optical frequency comb with a crystalline whispering gallery mode resonator. Physics Review Letters **101**, 093902 (2008).
26. Spencer, D. T. et al. An optical-frequency synthesizer using integrated photonics. Nature **557**, 81-85 (2018).
27. Marin-Palomo, P. et al. Microresonator-based solitons for massively parallel coherent optical communications. Nature **546**, 274 (2017).
28. Kues, M. et al. Quantum optical microcombs. Nature Photonics **13**, 170-179, doi:10.1038/s41566-019-0363-0 (2019).
29. Cole, D. C., Lamb, E. S., Del'Haye, P., Diddams, S. A. & Papp, S. B. Soliton crystals in Kerr resonators. Nature Photonics **11**, 671 (2017).
30. Stern, B., Ji, X., Okawachi, Y., Gaeta, A. L. & Lipson, M. Battery-operated integrated frequency comb generator. Nature **562**, 401 (2018).
31. Wu, J. *et al.* RF Photonics: An Optical Microcombs' Perspective. IEEE Journal of Selected Topics in Quantum Electronics **24**, 1-20 (2018).
32. Krizhevsky A, Sutskever I, Hinton G E. Imagenet classification with deep convolutional neural networks. Advances in neural information processing systems **25**, 1097-1105 (2012).
33. Metcalf, A. J. et al. Integrated line-by-line optical pulse shaper for high-fidelity and rapidly reconfigurable RF-filtering. Optics Express **24**, 23925-23940 (2016).
34. Sahin, E., Ooi, K., Png, C. & Tan, D. Large, scalable dispersion engineering using cladding-modulated Bragg gratings on a silicon chip. Applied Physics Letters **110**, 161113 (2017).
35. Wang, C. et al. Integrated lithium niobate electro-optic modulators operating at CMOS-compatible voltages. Nature **562**, 101 (2018).
36. Esteva, A. et al. Dermatologist-level classification of skin cancer with deep neural networks. Nature 542, 115-118 (2017).
37. Capper, D. et al. DNA methylation-based classification of central nervous system tumours. Nature **555**, 469, d oi:10.1038/nature26000
38. LeCun, Y., Bottou, L., Bengio, Y. & Haffner, P. Gradient-based learning applied to document recognition. Proceedings of the IEEE **86**, 2278-2324 (1998).
39. Bishop, C. M. Neural networks for pattern recognition. (Oxford university press, 1995).

**Funding:** This work was supported by the Australian Research Council Discovery Projects Program (No. DP150104327). R. M. acknowledges support by the Natural Sciences and Engineering Research Council of Canada (NSERC) through the Strategic, Discovery and Acceleration Grants Schemes, by the MESI PSR-SIIRI Initiative in Quebec, and by the Canada Research Chair Program. Brent E. Little was supported by the Strategic Priority Research Program of the Chinese Academy of Sciences, Grant No. XDB24030000. D.G.H was supported in part by the Australian Research Council grant FT104101104. R.M is affiliated with 7 as an adjoint faculty. **Author contributions:** X. X. conceived the idea and designed the project. X. X. and M. T. performed the experiments. X. X. analyzed the data, and performed the numerical simulations and the offline training. S.T.C. and B.E.L. designed and fabricated the integrated devices. B. C., A. B, T. N, R. M. and A. M. contributed to the development of the experiment and to the data analysis. X. X. and D. J. M. wrote manuscript. D. J. M. supervised the research. **Competing interests:** Authors declare no competing interests; and **Data and materials availability:** All data is available in the main text or the supplementary materials.

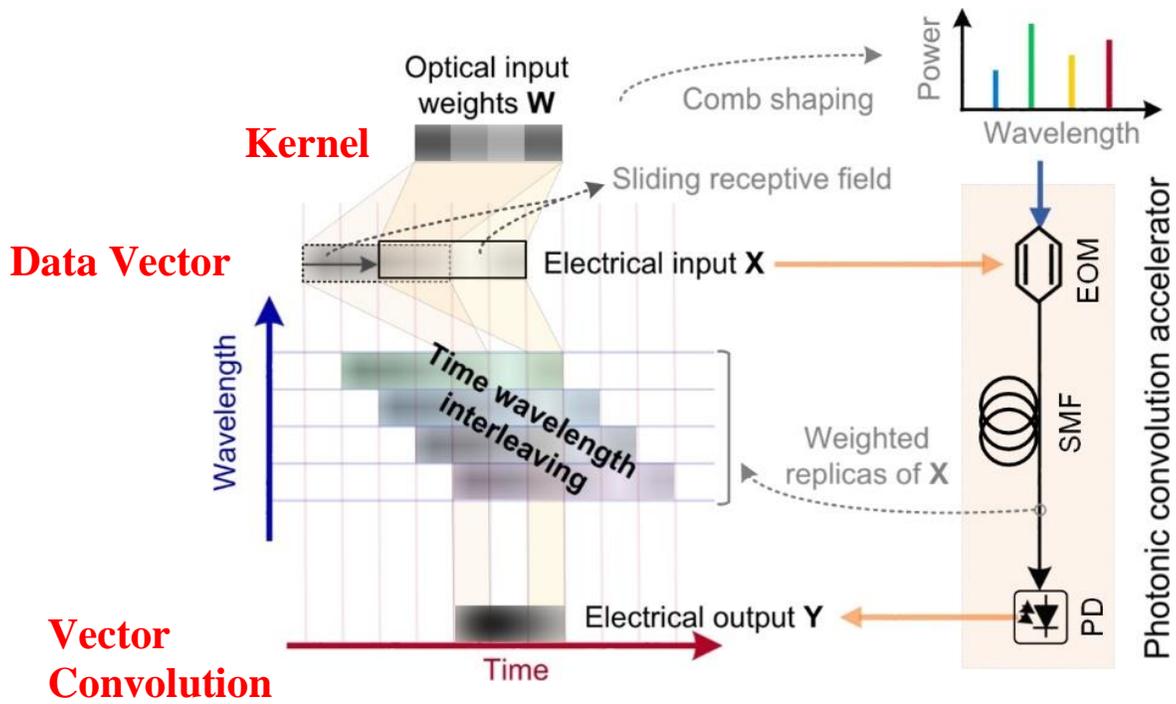

**Figure 1 | Operation principle of the Tera-FLOPS photonic convolution accelerator**. EOM: electro-optical Mach-Zehnder modulator. SMF: standard single mode fibre for telecommunications. PD: photodetector.

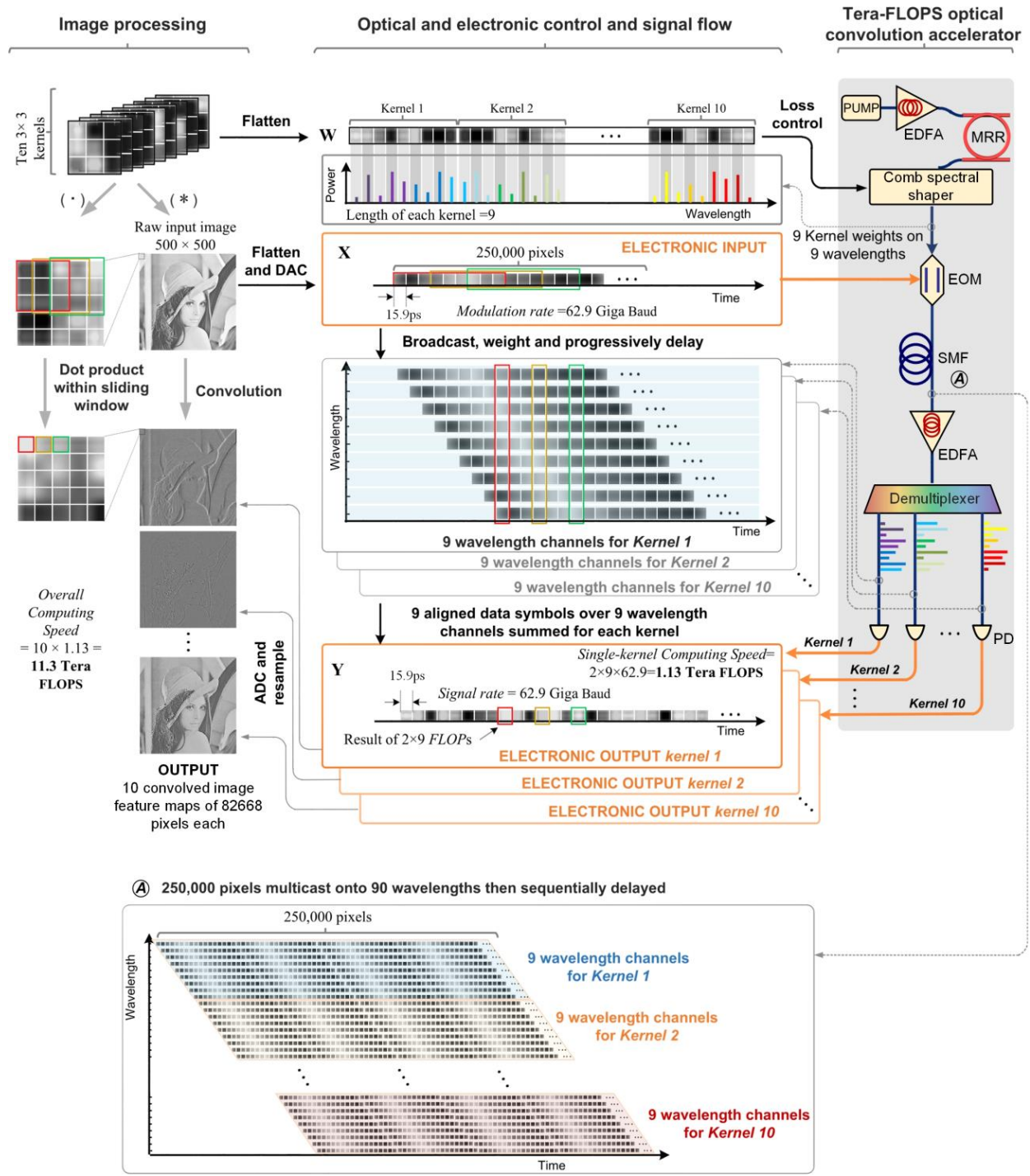

**Figure 2 | Image processing**, consisting of the experimental setup (right panel), the optical and electronic control and signal flow (middle panel), and the corresponding processing flow of the raw input image (left panel). CW pump: continues-wave pump laser. PC: polarization controller. EDFA: erbium doped fibre amplifier. MRR: micro-ring resonator. EOM: electro-optical Mach-Zehnder modulator. SMF: standard single mode fibre for telecommunications. PD: photodetector.

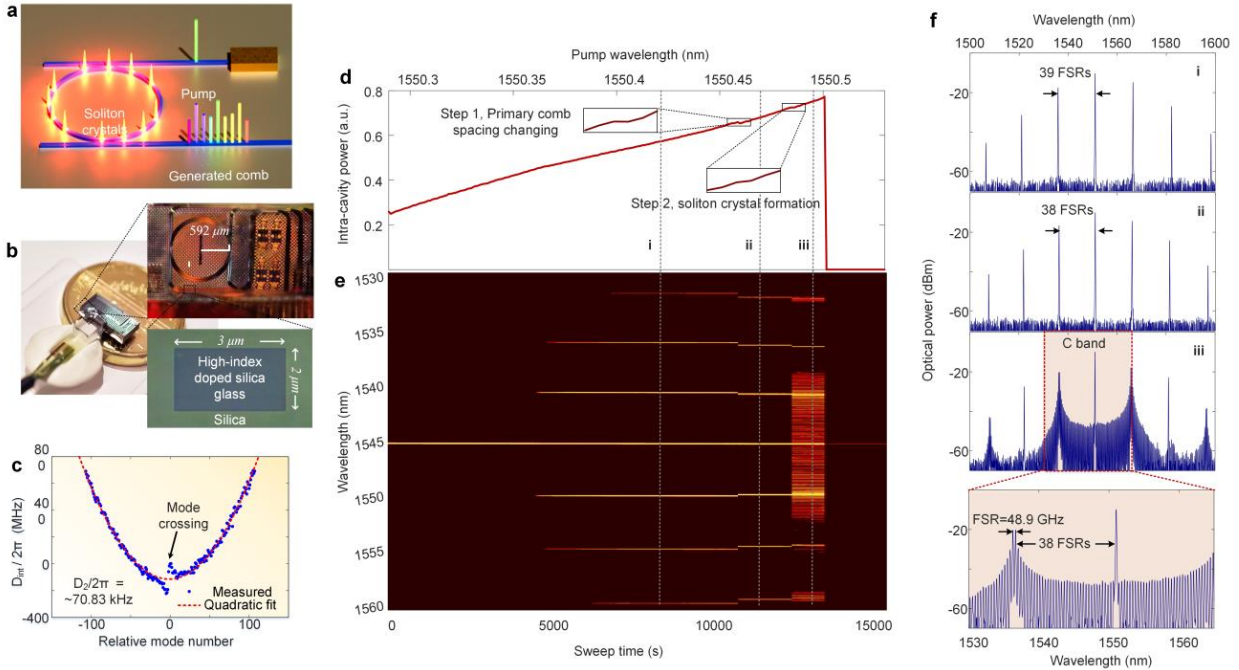

**Figure 3 | a.** Schematic diagram of the soliton crystal microcomb, generated by pumping an on-chip high-Q nonlinear micro-ring resonator with a continuous-wave laser. **b.** Image of the MRR and a scanning electron microscope image of the MRR's waveguide cross section. **c.** Measured dispersion of the MRR showing the mode crossing at ~1552 nm. **d.** Measured soliton crystal step of the intra-cavity power, and **e.** optical spectrum of the microcomb when sweeping the pump wavelength. **f.** Optical spectrum of the generated coherent microcomb at different pump detunings at a fixed power.

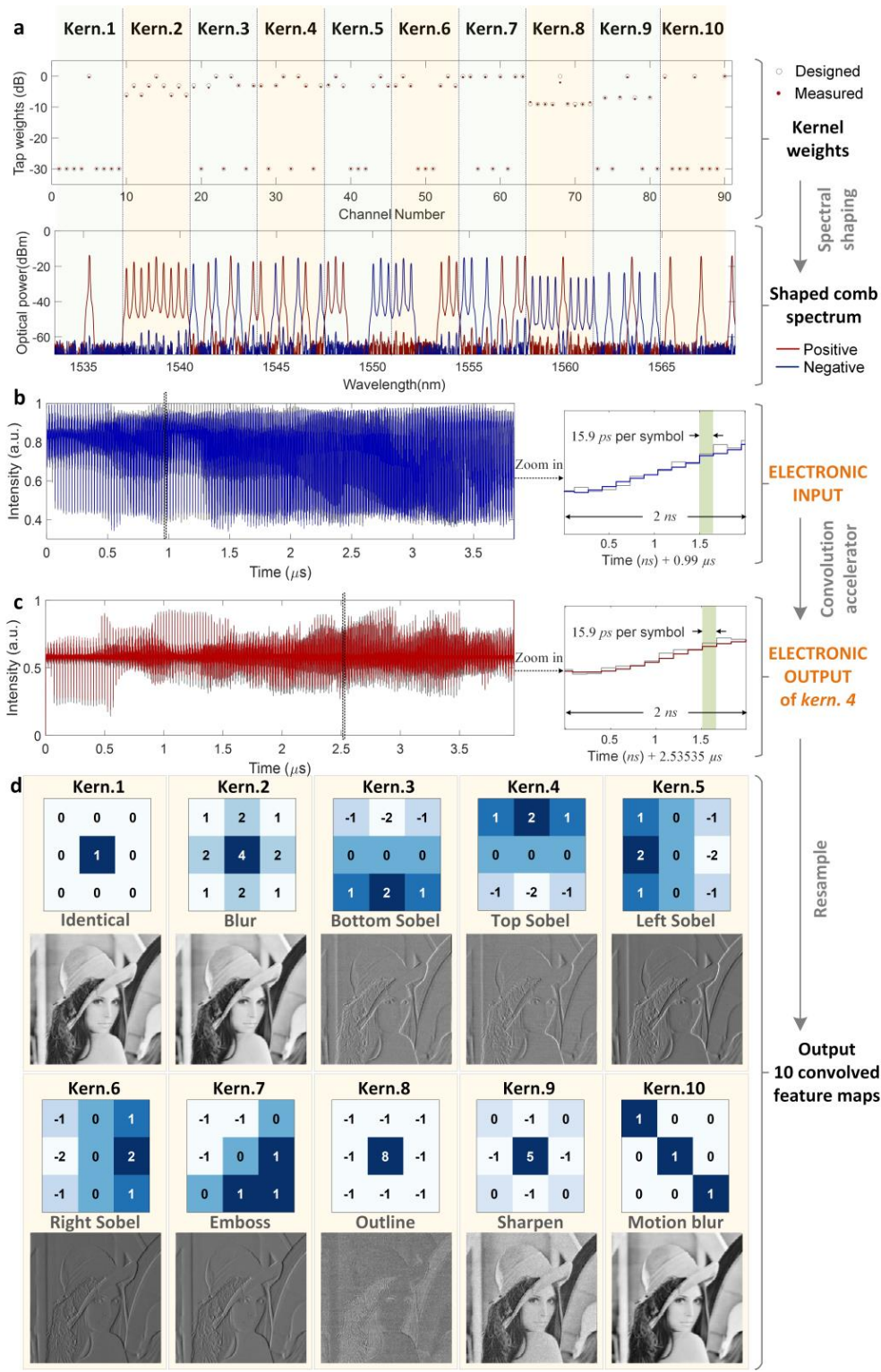

**Figure 4 | Experimental results of the image processing.** a. The kernel weights and the shaped microcomb's optical spectrum. b. The input electrical waveform of the image (the grey and blue lines show the ideal and experimentally generated waveforms, respectively). c. The convolved results of the fourth kernel that performs a top Sobel image (see Supplementary) processing function (the grey and red lines show the ideal and experimentally generated waveforms, respectively). d. The weight matrices of the kernels and corresponding recovered images.

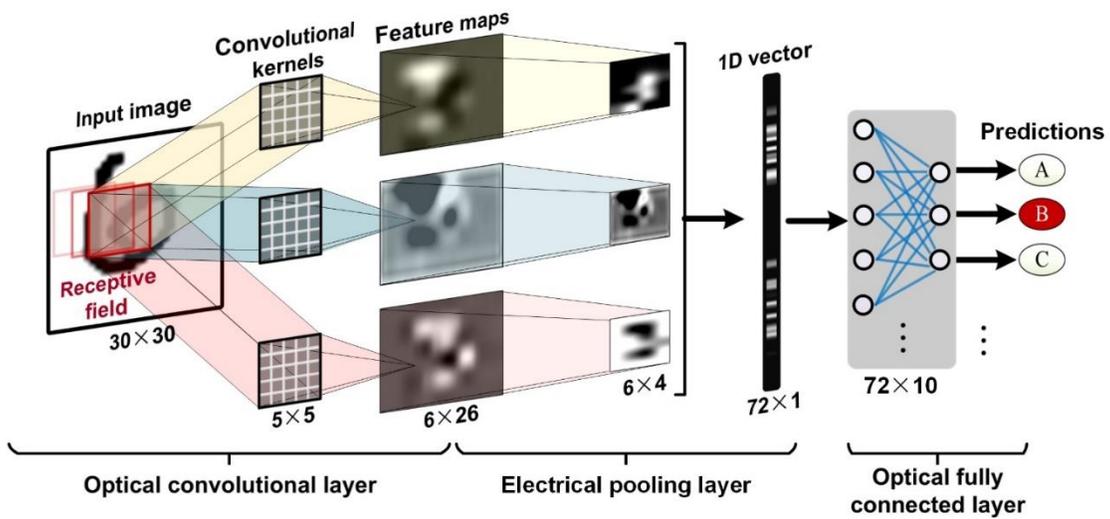

**Figure 5 | The architecture of the optical CNN**, including a convolutional layer, a pooling layer, and a fully connected layer.

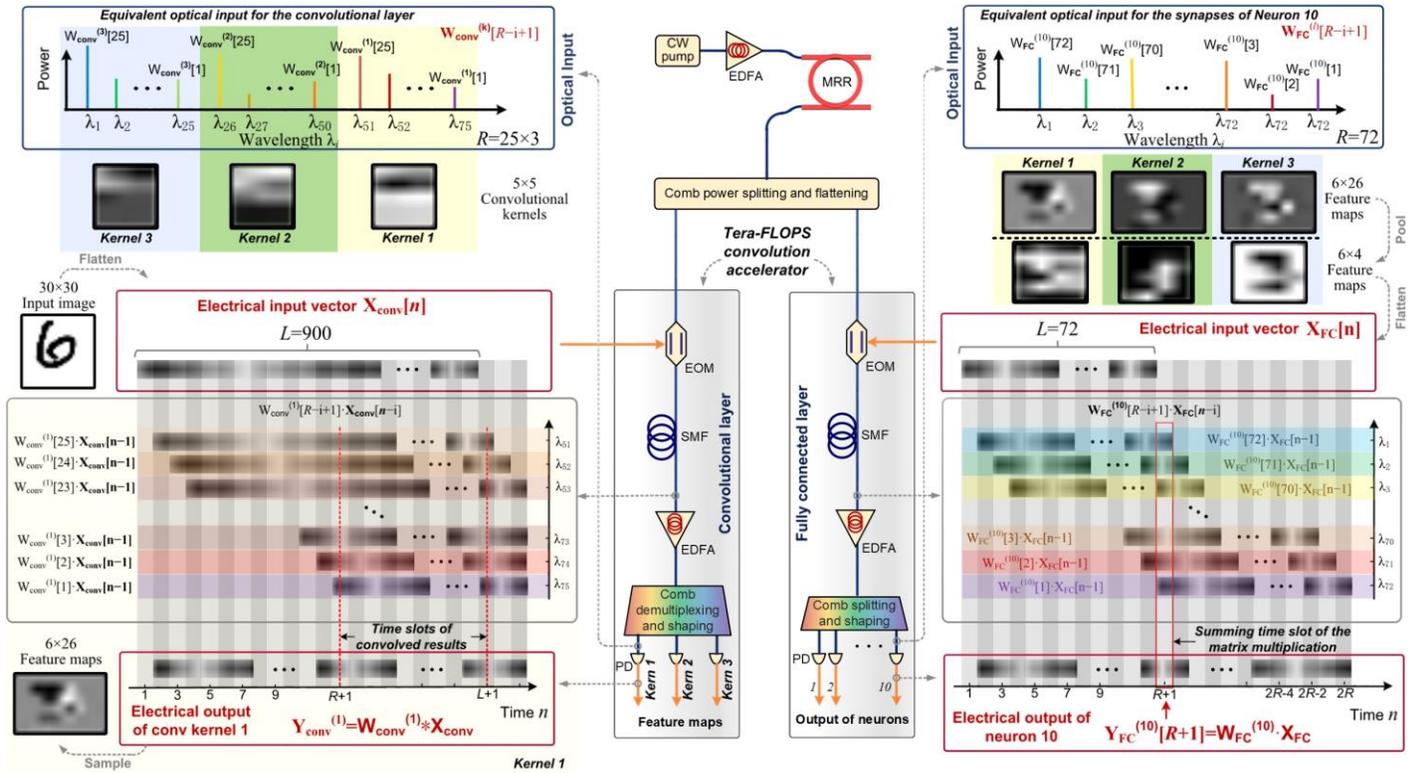

**Figure 6 | Experimental schematic of the optical CNN.** Left side is the input front end convolutional accelerator while the right side is the fully connected layer, both of which form the deep learning optical CNN. The microcomb source supplies the wavelengths for both the tera-FLOPS photonic convolution accelerator as well as the fully connected layer systems. The electronic digital signal processing (DSP) module used for sampling and pooling etc. is external to this structure.

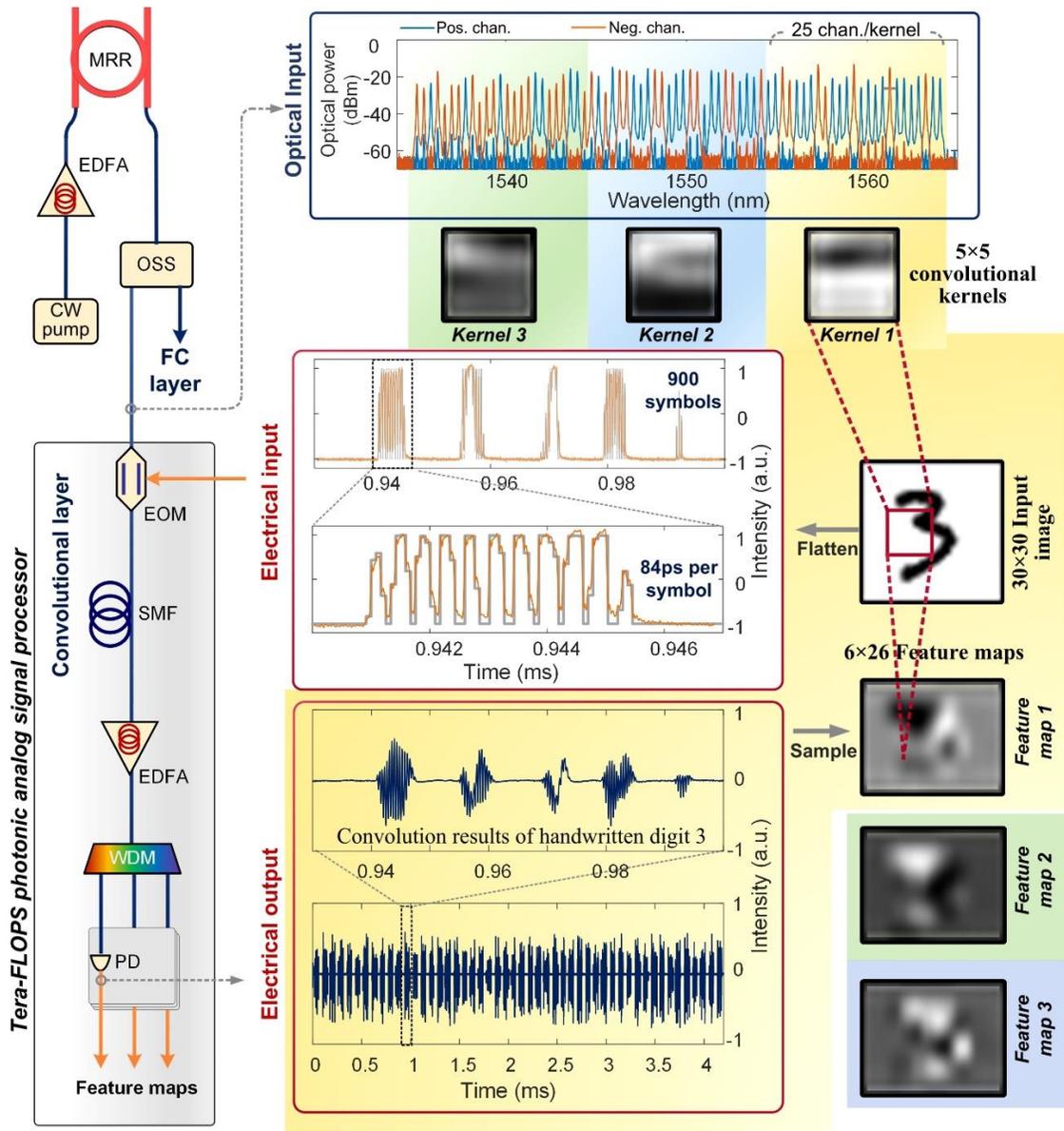

**Figure 7 | Convolutional layer. Architecture and experimental results.** The left panel shows the experimental setup. The right panel shows the experimental results of one of the convolutional kernels, showing the shaped microcomb's optical spectrum and the corresponding kernel weights (the blue and red lines denote the positive and negative synaptic weights, respectively), the input electrical waveform for the digit 3 (middle: the grey and yellow lines show the ideal and experimentally generated waveforms, respectively), the convolved results and the corresponding feature maps. CW pump: continuous-wave pump laser. PC: polarization controller. EDFA: erbium doped fibre amplifier. MRR: micro-ring resonator. OSS: optical spectral shaper. EOM: electro-optical Mach-Zehnder modulator. SMF: standard single mode telecommunications fibre. WDM: wavelength de-multiplexer. PD: photodetector. FC: fully connected.

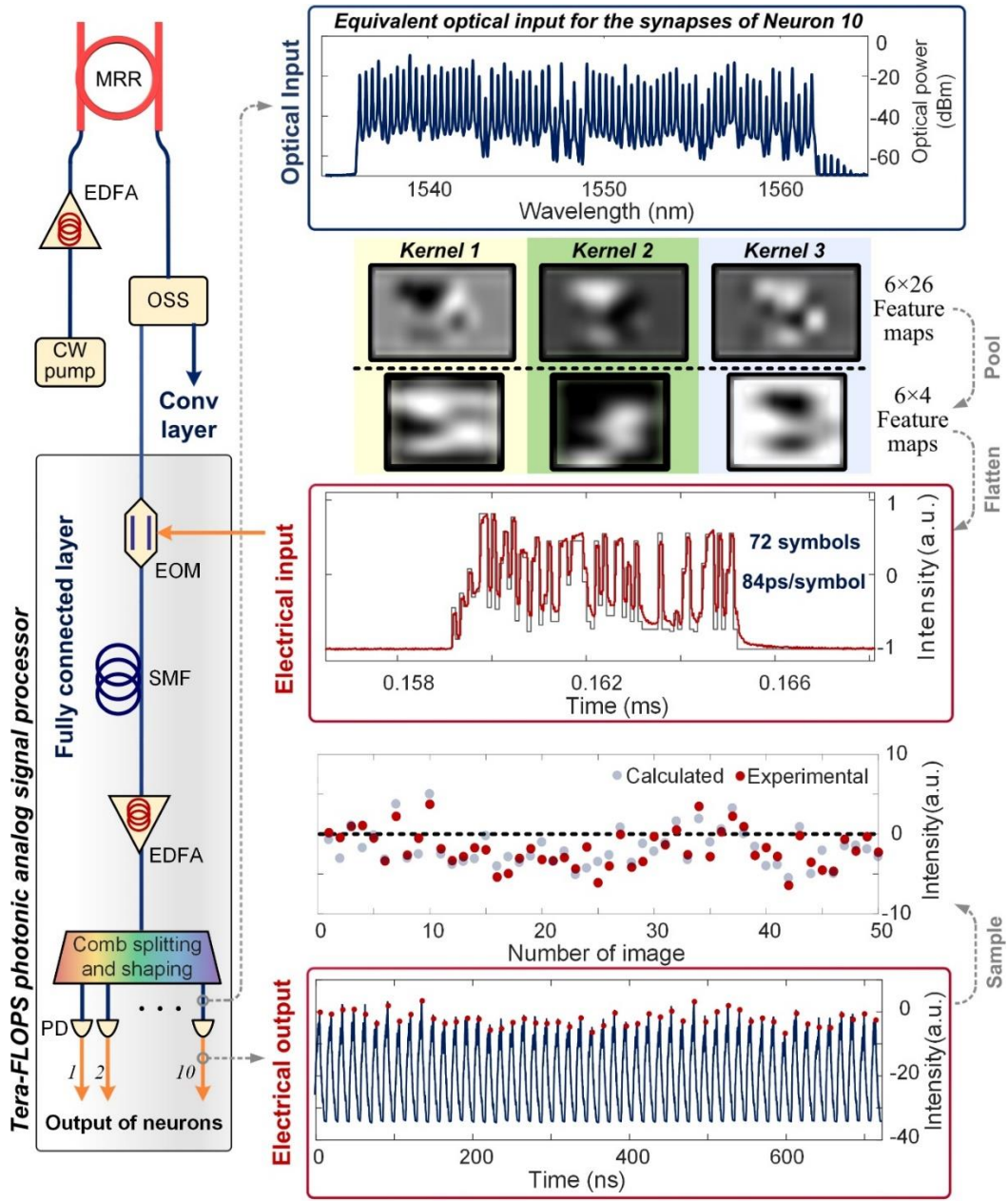

**Figure 8 | Fully connected layers. Architecture and experimental results.** The left panel depicts the experimental setup, similar to the convolutional layer. The right panel shows top: the experimental results for one output neuron, including the shaped comb spectrum; middle: the pooled feature maps of the digit "3" and the corresponding input electrical waveform (the grey and red lines illustrate the ideal and experimentally generated waveforms, respectively); and bottom: the output waveform of the neuron and sampled intensities.

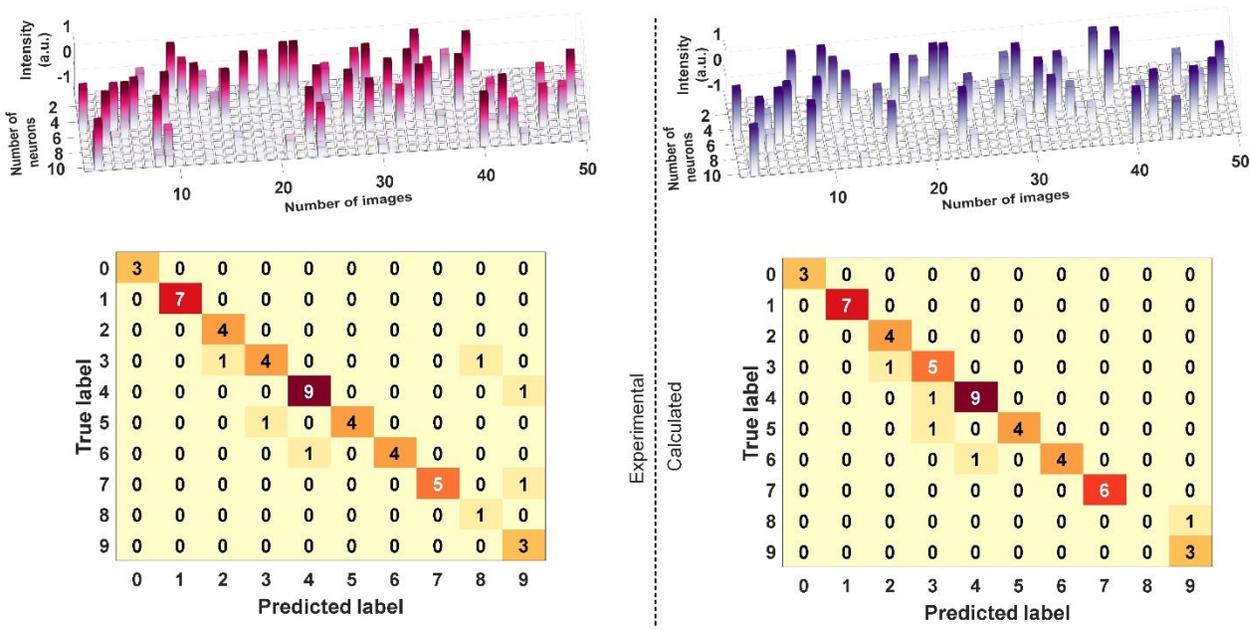

**Figure 9 | Experimental and theoretically calculated results for image recognition.** The upper figures show the sampled intensities of the ten output neurons at the fully connected layer, while the lower figures show the confusion matrices (see Supplementary Materials) with the darker colours indicating a higher recognition score.

## Methods

**Optical soliton crystal micro-comb**

Optical frequency combs, composed of discrete and equally spaced frequency lines, are extremely powerful tools for optical frequency metrology [32]. Micro-combs offer the full power of optical frequency combs, but in an integrated form with much smaller footprint [22-24]. They have enabled many breakthroughs through their ability to generate wideband low-noise optical frequency lines for high-resolution optical frequency synthesis [26], ultrahigh-capacity communications [27], complex quantum state generation [28], advanced microwave signal processing [31], and more.

In this work we use a particular class of microcomb termed soliton crystals. They were so-named because of their crystal-like profile in the angular domain of tightly packed self-localized pulses within micro-ring resonators [30]. They are naturally formed in micro-cavities with appropriate mode crossings, without the need for complex dynamic pumping and stabilization schemes (described by the Lugiato-Lefever equation [22]). They are characterized by distinctive 'fingerprint' optical spectra (Fig. 2f) which arise from spectral interference between the tightly packaged solitons circulating along the ring cavity. This category of soliton micro-comb features deterministic soliton formation originating from the mode crossing-induced background wave and the high intra-cavity power (the mode crossing is measured as in Fig. 2c). This in turn enables simple and reliable initiation via adiabatic pump wavelength sweeping [29] that can be achieved with manual detuning (the intracavity power during the pump sweeping is shown in Fig. 2d). The key to the ability to adiabatically sweep the pump lies in the fact that the intra-cavity power is over thirty times higher than single-soliton states (DKS), and very close to that of spatiotemporal chaotic states [22]. Thus, the soliton crystal displays much less thermal detuning or instability resulting from the 'soliton step' that makes resonant pumping of DKS states more challenging [22]. It is this combination of ease of generation and overall conversion efficiency that makes soliton crystals highly suited for demanding applications such as ONNs.

The coherent soliton crystal microcomb (Fig. 2) was generated by optical parametric oscillation in a single integrated micro-ring resonator (MRR). The MRR (Fig. 2b) was fabricated on a CMOS-compatible doped silica platform [22, 23], featuring a high Q factor of over 1.5 million and a radius of 592 μm, which corresponds to a low free spectral range of ~ 48.9 GHz [31]. The pump laser (Yenista Tunics – 100S-HP) was boosted by an optical amplifier (Pritel PMFA-37) to initiate the parametric oscillation. The soliton crystal microcomb provided over 90 channels over the telecommunications C-band (1540-1570 nm), offering adiabatically generated low-noise frequency comb lines with a small footprint of < 1 $mm^2$ and potentially low power consumption (>100 mW using the technique in [30]).

**Evaluation of the computing performance**

Since there are no common standards in the literature for classifying and quantifying the computing speed and processing power of ONNs, we explicitly outline the performance definitions that we use in characterizing our performance. We follow the approach that is widely used to evaluate electronic micro-processors. The computing power of the convolution accelerator—closely related to the operation bandwidth—is denoted as the *throughput*, which is the number of operations performed within a certain period. Considering that in our system the input data and weight vectors originate from different paths and are interleaved in different dimensions (time, wavelength, and space), we use the temporal sequence at the electrical output port to define the *throughput* in a more straightforward manner.

At the electrical output port, the output waveform has $L+R-1$ symbols in total ($L$ and $R$ are the lengths of the input data vector and the kernel weight vector, respectively), among which $L-R+1$ symbols are the convolution results. Further, each output symbol is the calculated outcome of $R$ multiply-and-accumulate operations or $2R$ FLOPS, with a symbol duration $\tau$ given by that of the input waveform symbols. Thus, considering that $L$ is generally much larger than $R$ in practical convolutional neural networks, the term $(L-R+1)/(L+R-1)$ would not affect the vector computing speed, or *throughput*, which (in FLOPS) is given by

$$\frac{2R}{\tau} \cdot \frac{L-R+1}{L+R-1} \approx \frac{2R}{\tau} \qquad (1)$$

As such, the computing speed of the vector convolutional accelerator demonstrated here is $2 \times 9 \times 62.9 \times 10 = 11.321$ Tera-FLOPS for ten parallel convolutional kernels).

We note that when processing data in the form of vectors, such as audio speech, the effective computing speed of the accelerator would be the same as the vector computing speed $2R/\tau$. Yet when processing data in the form of matrices, such as for images, we must account for the overhead on the effective computing speed brought about by the matrix-to-vector flattening process. The overhead is directly related to the width of the convolutional kernels, for example, with 3-by-3 kernels, the effective computing speed would be $\sim 1/3 * 2R/\tau$, which, however, we note still is in the ultrafast (TeraFLOP) regime due to the high parallelism brought about by the time-wavelength interleaving technique.

For the convolutional accelerator, the output waveform of each kernel (with a length of $L-R+1=250,000-9+1=249,992$) contains $166 \times 498 = 82,668$ useful symbols that are sampled out to form the feature map, while the rest of the symbols are discarded. As such, the effective matrix convolution speed for the experimentally performed task is slower than the vector computing speed of the convolution accelerator by the overhead factor of 3, and so the net speed then becomes $11.321 \times 82,668/249,991 = 11.321 \times 33.07\% = 3.7437$ Tera-FLOPS.

For the deep CNN the convolutional accelerator front end layer has a vector computing speed of $2 \times 25 \times 11.9 \times 3 = 1.785$ Tera-FLOPS while the matrix convolution speed for 5x5 kernels is $1.785 \times 6 \times 26/(900-25+1) = 317.9$ Giga-FLOPS. For the fully connected layer of the deep CNN, according to Eq. (4), the output waveform of each neuron would have a length of $2R-1$, while the useful (relevant output) symbol would be the one locating at $R+1$, which is also the result of $2R$ operations. As such, the computing speed of the fully connected layer would be $2R / (\tau*(2R-1))$ per neuron. With $R = 72$ during the experiment and ten neurons simultaneous operating, the effective computing speed of the matrix multiplication would be $2R / (\tau*(2R-1)) \times 10 = 2 \times 72 / (84ps* (2 \times 72-1)) = 119.83$ Giga-FLOPS.

In addition, the intensity resolution (i.e., the bit-resolution for digital systems) for analog ONNs is mainly limited by the signal-to-noise ratio (SNR). To achieve 8-bit resolution, the SNR of the system needs to reach over $20 \cdot \log10(2^8) = 48$ dB. This is within the capability of our accelerator and so our system speed in Terabits/s is simply our speed in FLOPs times 8 – ie., not reduced by our OSNR.

**Experiment**

To achieve the designed kernel weights, the generated microcomb was shaped in power using liquid crystal on silicon based spectral shapers (Finisar WaveShaper 4000S). We used two

WaveShapers in the experiments - the first was used to flatten the microcomb spectrum while the precise comb power shaping required to imprint the kernel weights was performed by the second, located just before the photo-detection. A feedback loop was employed to improve the accuracy of comb shaping, where the error signal was generated by first measuring the impulse response of the system with a Gaussian pulse input and comparing it with the ideal channel weights. (Figure S6 and S7 show the shaped impulse response for the convolutional layer and the fully connected layer of the CNN).

The electrical input data was temporally encoded by an arbitrary waveform generator (Keysight M8195A) and then multicast onto the wavelength channels via a 40 GHz intensity modulator (iXblue). For the 500×500 image processing, we used sample points at a rate of 62.9 Giga samples/s to form the input symbols. We then employed a 2.2 km length of dispersive fibre that profiided a progressive delay of 15.9 ps/channel, precisely matched to the input baud rate. For the convolutional layer of the CNN, we used 5 sample points at 59.421642 Giga Samples/s to form each single symbol of the input waveform, which also matched with the progressive time delay (84 ps) of the 13km dispersive fibre (the generated electronic waveforms for 50 images are shown as Fig. S8 and S9, which served as the electrical input signal for the convolutional and fully connected layers, respectively).

For the convolutional accelerator in both experiments - the 500×500 image processing experiment and the convolutional layer of the CNN - the second Waveshaper simultaneously shaped and de-multiplexed the wavelength channels into separate spatial ports according to the configuration of the convolutional kernels. As for the fully connected layer, the second Waveshaper simultaneously performed the shaping and power splitting (instead of de-multiplexing) for the ten output neurons. Here, we note that the de-multiplexed or power-split spatial ports were sequentially detected and measured. However, these two functions could readily be achieved in parallel with a commercially available 20-port optical spectral shaper (WaveShaper 16000S, Finisar) and multiple photodetectors.

The negative channel weights were achieved using two methods. For the 500×500 image processing experiment and the convolutional layer of the CNN, the wavelength channels of each kernel were separated into two spatial outputs by the WaveShaper according to the signs of the kernel weights, and then detected by a balanced photodetector (Finisar XPDV2020). Conversely, for the fully connected layer the weights were encoded in the symbols of the input electrical waveform during the electrical digital processing stage. Incidentally, we demonstrate the possibility using of different methods to impart negative weights, both of which work in the experiments.

Finally, the electrical output waveform was sampled and digitized by a high-speed oscilloscope (Keysight DSOZ504A, 80 Giga Symbols/s) to extract the final convolved output.

In the CNN, the extracted outputs of the convolution accelerator were further processed digitally, including rescaling to exclude the loss of the photonic link via a reference bit, and then mapped onto a certain range using a nonlinear *tanh* function. The pooling layer's functions were also implemented digitally, following the algorithm introduced in the network model.

The residual discrepancy or inaccuracy in our work for both the recognition and convolving functions, as compared to the numerical calculations, was due to the deterioration of the input waveform caused by intrinsic limitations in the performance of the electrical arbitrary waveform generator. Addressing this would readily lead to a higher degree of accuracy (i.e., closer agreement with the numerical calculations).